%% file: root.tex
\documentclass[letterpaper, 10 pt, conference]{ieeeconf}  

\IEEEoverridecommandlockouts                              

\usepackage{graphics} 
\usepackage{graphicx}
\usepackage{balance}
\usepackage{xcolor}

\usepackage{amsmath} 
\usepackage{amssymb}  
\usepackage{float}

\usepackage[ruled,vlined]{algorithm2e}
\usepackage{booktabs}

\title{\LARGE \bf
Learning Multimodal Contact-Rich Skills from Demonstrations Without Reward Engineering}

\author{Mythra V. Balakuntala$^{1}$, Upinder Kaur$^{1}$, Xin Ma$^{*1}$, Juan Wachs$^{2}$,  Richard M. Voyles$^{1}$
\thanks{This work was supported by the NSF Center on RObots and SEnsors for HUman well-Being (RoSe-HuB) under grant CNS-1439717 and by the Office of the Assistant Secretary of Defense for Health Affairs under Award No. W81XWH-18-1-0769.}
\thanks{$^{*}$Corresponding author, $^{1}$School of Engineering Technology, $^{2}$School of Industrial Engineering, 
        Purdue University, IN 47907, USA
        {\tt\small mbalakun@purdue.edu, kauru@purdue.edu, maxin1988maxin@gmail.com, rvoyles@purdue.edu, jpwachs@purdue.edu}}%
        } 
\begin{document}

\maketitle
\thispagestyle{empty}
\pagestyle{empty}
\begin{abstract}
Everyday contact-rich tasks, such as peeling, cleaning, and writing, demand multimodal perception for effective and precise task execution. However, these present a novel challenge to robots as they lack the ability to combine these multimodal stimuli for performing contact-rich tasks. Learning-based methods have attempted to model multi-modal contact-rich tasks, but they often require extensive training examples and task-specific reward functions which limits their practicality and scope. Hence, we propose a generalizable model-free learning-from-demonstration framework for robots to learn contact-rich skills without explicit reward engineering. We present a novel multi-modal sensor data representation which improves the learning performance for contact-rich skills. We performed training and experiments using the real-life Sawyer robot for three everyday contact-rich skills -- cleaning, writing, and peeling. Notably, the framework achieves a success rate of 100\% for the peeling and writing skill, and 80\% for the cleaning skill. Hence, this skill learning framework can be extended for learning other physical manipulation skills.

\end{abstract}
\section{INTRODUCTION}
Seemingly trivial contact-rich skills, such as peeling a cucumber or cleaning a surface, demand seamless coordination among eyes and hands in humans. While eyes provide object pose, environment state, and visual feedback; haptic feedback from hands provides contact information between objects, relative localization, and alternative control in case of loss of visual feedback. Visual and tactile modalities -- each help resolve ambiguities in the other -- thereby deliver an overall accurate perception of dynamic environments \cite{blake2004neural}. This synergistic interaction between the two modalities of sensing is essential for contact-rich tasks, especially in the case of robots. 

Robots, aiming to achieve human-level perception and proprioception for a variety of tasks, need the ability to better understand and model their interactions with the environment. While numerous published works in robotics have recognized the utility of a multi-modal system \cite{song2014automated,jmromano, levine2016end}, their implementation has been limited to task-specific solutions. These works have used extensive task-specific feature engineering coupled with prior knowledge of task execution in model-based methods, which fairly limits their generalized use. Subsequently, model-free methods have been introduced which attempt to learn multi-modal contact-rich skills directly from observations. These methods are preferred due to the low cost of implementation, re-programmability, and the ability to model complex tasks previously out of the scope of model-based methods.   

Learning from Demonstration (LfD) is a popular approach which includes model-free methods to ``program" a robot by leveraging demonstrations of the tasks \cite{argall2009survey,calinon2007teacher, calinon2007incremental}. Sequences of demonstrations allow the robots to comprehend the objective, action space, and the manipulation required for completing the task \cite{balakuntala2019extending}. Architectures such as Gaussian Mixture Models, Hidden Markov Models, Deep Neural Networks, and Dynamic Motion Primitives (DMP) have been explored for learning from demonstrations \cite{billard2016learning,argall2009survey}. These methods can generalize using extensive demonstrations of kinematic states for tasks such as grasping and pushing simple objects. However, not only is getting large kinematic samples a limitation of these methods, but they are also unable to handle perturbations in the workspace \cite{finn2017one,finn2017deep}. 
\begin{figure}[t]
    \centering
    \includegraphics[width=0.46\textwidth]{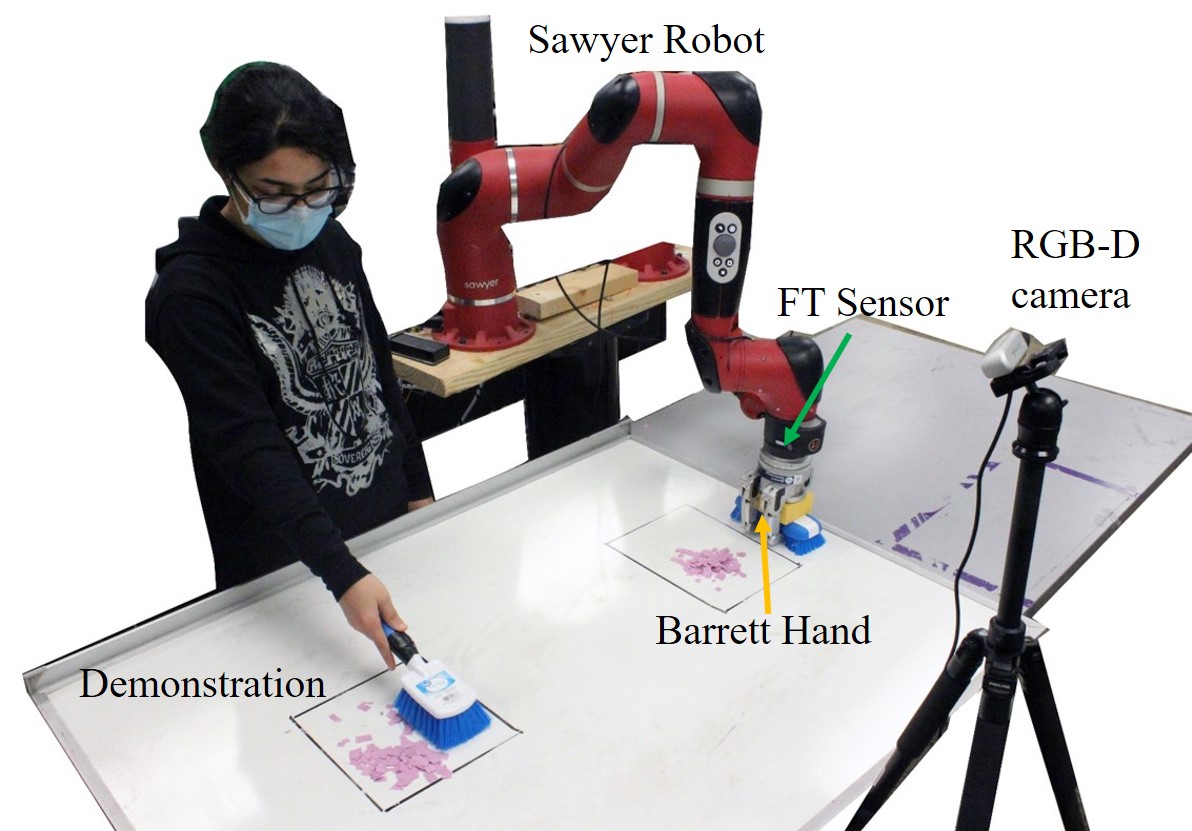}
    \caption{The task setup with human demonstrations to the robot for cleaning}
    \label{fig:setup}
\end{figure}

Reinforcement learning-based methods solve some of these issues, as they do not require explicit task specification, and can handle greater perturbations, however, they need well-tuned reward functions \cite{billard2016learning}. Inverse reinforcement learning overcomes this limitation by implicitly inferring rewards from demonstrations \cite{billard2016learning,abbeel2004apprenticeship, avi2019rss}. Yet, most of these methods are modeled in closed-loop systems around a single modality, majorly visual feedback \cite{lee2019making, chebotar2017path,finn2017deep,zhu2018reinforcement}. Using kinesthetic interfaces for teaching by demonstration enables learning of contact-intensive tasks by guided demonstrations from the user \cite{kalakrishnan2011learning,lee2015learning}. However, such interfaces not only limit the domain of tasks that can be taught but also puts the onus of effective learning on the guiding expertise of the demonstrator. Moreover, teaching movements and skills by kinesthetic learning demands long guided intervals that are cost and resource intensive. 

Contact-rich skills, such as peeling and writing, cannot be accomplished with just a single modality as both force and visual feedback is necessary for optimum execution. Few prior approaches have exploited the synergy between vision and touch for contact-rich tasks \cite{lee2019making}. Methods combining vision and touch have been employed to increase the stability of grasp, or learn shape completion for grasping \cite{bekiroglu2011learning,calandra2018more}. Further, other works utilize multiple modalities for a manipulation task, however, they needed task-specific strategies or a task-specific specified manipulation graph \cite{watkins2019multi,kappler2015data,abu2015adaptation}. 

In this work, we leverage multiple sensor modalities to learn contact-rich skills without task-specific reward engineering. Learning contact-rich skills is a significant step for robots to operate in unstructured environments. However, designing reward functions to learn such contact-rich skills is not trivial or in some cases infeasible. In this framework, we infer the reward from the goal states obtained from expert user demonstrations of the skill. The demonstrations are semantically segmented to identify human-object interactions and the goal states. Additionally, the interaction information is used to define the action space for the policy. A representation for the state is developed from the multi-sensor data for the policy, which can improve the learning performance for contact-rich skills. To demonstrate the practicality of this framework, we implement it using the Sawyer\texttrademark  robot. Three different contact-rich tasks (writing, cleaning, and peeling) requiring varying force, contact, and position interactions are demonstrated for testing the performance of the proposed framework.
The contributions of this work are:
\begin{itemize}
    \item A framework to learn contact-rich skill models from a few demonstrations without reward engineering, which can remove the need for manual engineering of rewards. 
    \item A multi-modal sensor data representation for learning contact-rich skills, which improves the overall execution performance.
    \item Evaluation of the framework on three varied real-life contact-rich skills -- cleaning, writing, and peeling.
\end{itemize}

This paper is organized as follows: Section II describes the problem statement and the overall LfD framework. Section III describes the multi-modal skill learning framework and the sub-components of this framework, such as the input semantic segmentation and the execution controller. The experiments and results are shown in Section IV for the three experiments with the conclusion presented in Section V along with the discussion of future work. 

\section{Problem Statement and LfD Framework}
\subsection{Problem Statement}
Teaching a robot contact-rich manipulation for a variety of tasks is a non-trivial endeavor. Contact-rich tasks entail learning precise models of direct low-level control to reduce contact forces and ensure stability for smooth task execution \cite{beltran}. The first step in this is for the robot to be able to comprehend the multi-modal stimuli of the the skill being taught or executed. Hence, just position-based control is insufficient; therefore, we use both position and force control. The input multi-sensory data, consisting of visual demonstration, depth inputs, force and tactile information, is first learned by various networks. These learned representations update the skill modeling policy.

The model executes in a finite bounded horizon, with a Markov decision process $\mathcal{M}$, with state space $\mathcal{S}$ and action space $\mathcal{A}$. For a horizon $T$, the state transitions according to the dynamics $\mathcal{T}:\mathcal{S} \times \mathcal{A} \longrightarrow \mathcal{S}$. A learning policy $\pi(a|s)$ represents the probability of taking the action $a$ given a state $s$. The cumulative expected reward over horizon $T$ is given by \eqref{eq:1}. 
\begin{equation}
    J(\pi) = \mathbb{E}_{\pi} \left[\sum_{t=1}^{T} \phi(s_{t},a_{t})\right]
\label{eq:1}
\end{equation}
This reward is limited in the range $\phi:\mathcal{S}\times\mathcal{A}  \to \left[ -1, 0\right]\bigcup \{1\}$, where the negative rewards penalize the policy for sub-optimal convergence, and $1$ being the maximum reward given on optimal convergence. The optimal stochastic policy $\pi^{*}$, for the model is achieved by maximizing the cumulative expected reward $J(\pi)$.
\begin{figure}[t]
    \centering
    \includegraphics[width=0.47\textwidth]{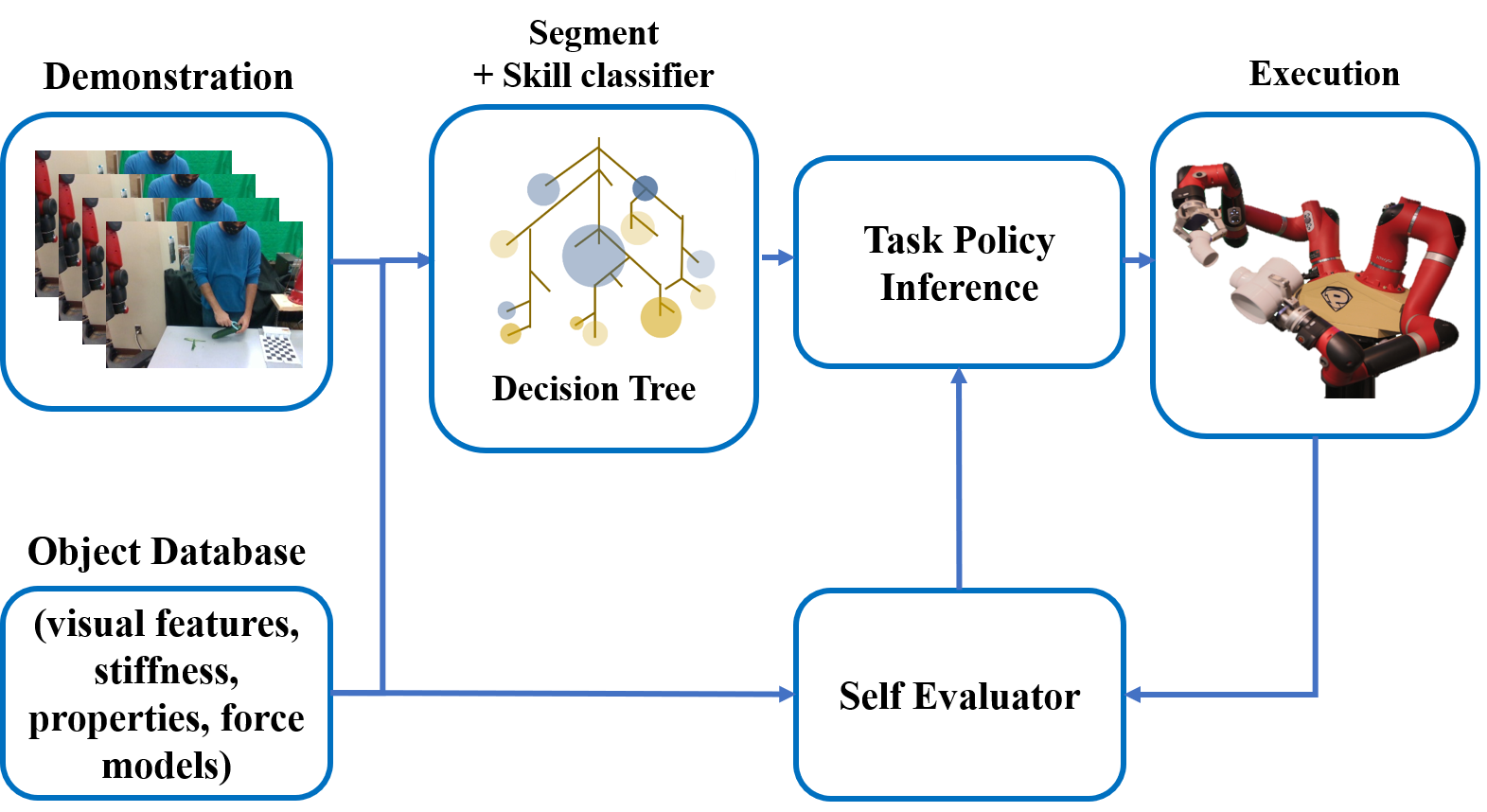}
    \caption{The Learning from Demonstration (LfD) Framework}
    \label{fig:lfd}
\end{figure}

\subsection{LfD Framework}
Unlike traditional Learning from Demonstration (LfD) approaches which rely on large number of demonstrations to learn a task, in our previous work, we presented a novel one-shot learning from demonstration approach, augmented by coaching, to learn tasks from one expert demonstration \cite{balakuntala2019extending}, as shown in Figure \ref{fig:lfd}. The demonstration is automatically segmented into a sequence of parametrized \textit{a priori} skills using a decision tree classifier. A self evaluator updates the skill paramters based on the execution performance to refine the task performance, locally optimizing cumulative performance. The \textit{a priori} skills are modelled controllers similar to DMPs \cite{ijspeert2013dynamical}. The current paper proposes a method to learn these contact-rich skills from expert demonstrations,  thereby eliminating the need for modelling the controllers.  

\begin{figure}[t]
    \centering
    \includegraphics[width=0.46\textwidth]{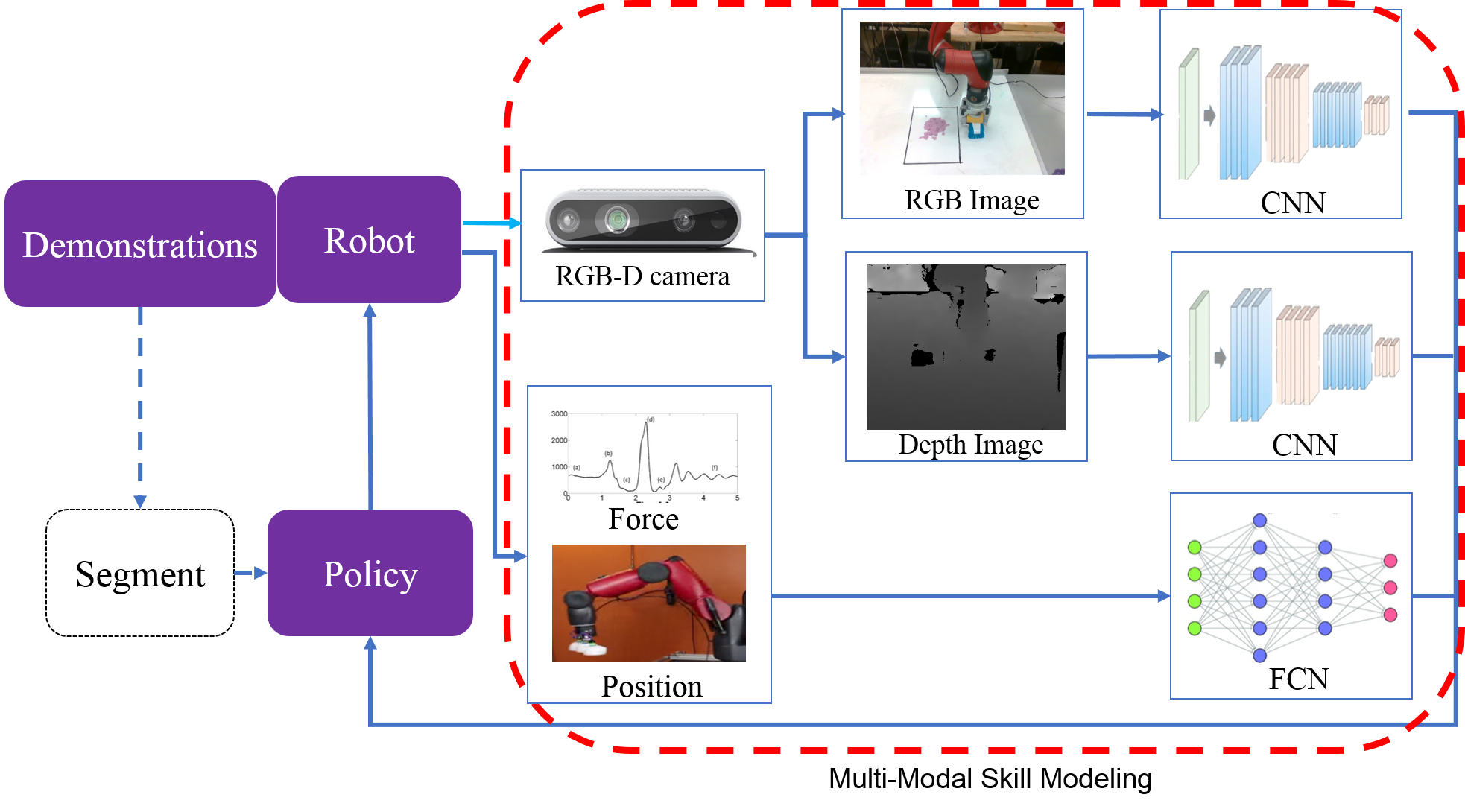}
    \caption{The skill learning framework}
    \label{fig:skill_lrn}
\end{figure}

\input{skill_framework.tex}

\input{experiments.tex}
\vspace{-0.15in}
\section{Conclusion and Discussion}
In this paper, we propose a novel framework to learn policies for contact-rich skills from few demonstrations without reward engineering. In the framework, the demonstrations are semantically segmented to identify human-object interactions and goal states. Subsequently, the reward can be learned from goals identified in the demonstrations, which removes the need for manual engineering of reward. Additionally, the interaction modes are used to define the action space for the policy. A multi-modal representation for the state is developed from the sensor data for the policy, which improves the learning performance for contact-rich skills. In the experiments, three different contact-rich skills requiring varying force and position interactions are demonstrated to evaluate performance of the proposed framework. Results show that the skill learning framework achieves a success rate of 100\% on writing and peeling skills and 80\% for the cleaning skill. In the future, this work will be extended to other manipulation skills with physical interaction such as cutting and scooping. 

\bibliographystyle{IEEEtran}
\balance
\bibliography{refs}
\end{document}

%% file: skill_framework.tex
\section{Multi-Modal Skill Learning Framework}
The proposed skill learning framework is designed to learn  contact-rich skills from few visual demonstration without requiring an explicitly defined reward. One drawback of using RL agents for real-world skills is the large exploration space and the need for reward shaping. To overcome this issue, we propose to use videos of expert human demonstrations to infer the action space and the reward. State representations which capture action relevant information can benefit from learning control behaviors \cite{lee2019making}. For contact-rich tasks, such representations should include force or contact information. We propose such a representation combining the data from multiple sensing modalities. The robot leverages this representation and the learnt reward to learn the execution policy for the skill. The different components of the skill learning framework are shown in Figure. \ref{fig:skill_lrn}. 

\subsection{Input Semantic Segmentation}\label{sec:seg}
The input for learning these skills is a set of few visual  human demonstrations. Identifying human-object interactions is a significant step to understand goal or intent of a skill \cite{gkioxari2018detecting}. Here, the frames that have such  human-object interactions are referred to as physical interaction keypoints (PIKs). The PIKs provide information about the goal states of objects. 

Instead of training an end-to-end detector for the PIKs,  existing object detection or segmentation  networks are utilized to generate features. Paramount to detecting physical interactions is identifying the hand and the surfaces of the objects in the scene. The hand pose and the object instance segmentation features are used as features to train a network to detect interaction between the hands and any object in each frame. The hand pose, interaction mode and hand motion can be used to infer whether the user is performing position or force control along a particular axis. This information can be leveraged to generate the action space for the skill policy. 

The hand pose (6-D) and contours are obtained using a YoloV3 \cite{redmon2018yolov3} detector. The object surfaces are obtained using instance segmentation from MaskRCNN \cite{he2017mask}. Features are generated from the hand pose and the instance segments to identify the PIK. The segmentation mask of dimension $128\times128$ is passed thorough a 4-layer convolutional network followed by a 3-layer multilayer perceptron (MLP) to result in 24 dimension object features. These features are concatenated with the hand poses of dimension $12$ to result in a $36$ dimension input to a 2-layer MLP network. The network outputs two binary values, indicating if the left and right hand are interacting with any object respectively.

The same $36$ features are used as input to another 2-layer MLP which outputs the class of the object each hand is interacting with. The frames where the interaction condition switches are denoted as PIKs. The first PIK denotes the start of the skill and all further PIKs are assumed to be positive goal states. 

The motion of the hand, the hand pose, and the interaction  condition in a moving window of 10 frames is used to form the action space. The action space for contact tasks is formulated as a tuple $(A_x,A_F)$. $A_x$ denotes the domain for kinematic exploration using position changes. For a surface contact task this will be all degrees of freedom (DoF) other than the contact direction. $A_F$ indicates the force along the contact direction. Clearly, $A_F$ and $A_x$ are along perpendicular directions in space. The goal examples and the action space obtained from segmentation are used for the policy training.
\begin{figure}[t]
    \centering
    \includegraphics[scale=0.25]{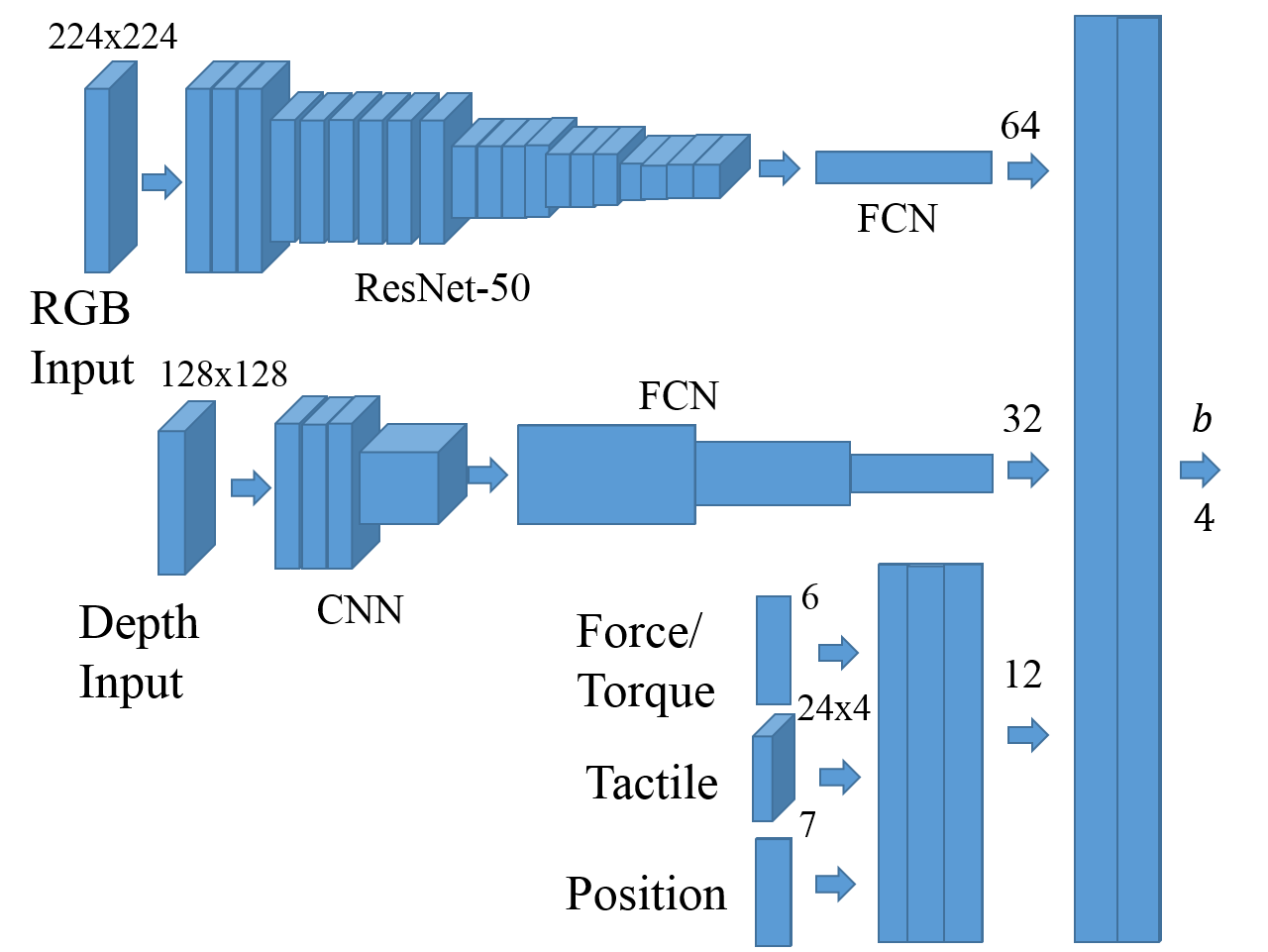}
    \caption{The skill learning framework}
    \label{fig:multimodal_rep}
\end{figure}

\subsection{Multi-Modal Sensor Representation}
Learning contact-rich skills require inputs from force, position and visual sensors to effectively understand the environment state. Combining data from multiple sensors can effectively learn tasks such as grasping or peg insertion \cite{calandra2018more, lee2019making}. Predictive representations for state based on multi sensor data can capture action-relevant information \cite{lesort2018state}. Inspired by this we propose a multi-modal representation to capture contact conditions, the type of contact and occlusion to camera based on the multiple sensor inputs.

Our model leverages deep networks to learn a representation for the state using the information from 5 different inputs. The inputs include an RGB image $I_c$, a depth image $I_d$, forces and torques in 6 DoF, tactile measurements and joint positions. Fig. \ref{fig:multimodal_rep}. The state representation consists of four binary decision values - existence of contact, type of contact, presence of tactile forces, and binary decision on camera occlusion.

The RGB and depth images are obtained from the Intel Realsense D435i. The images have a size of $640 \times 480$. The RGB images are rescaled to $224 \times 224$ and are passed through a ResNet-50 \cite{he2016deep} backend and one fully connected layer to provide output features of dimension 64. The depth images capture important scene information regarding occlusions and object positions. The depth images are rescaled to $128\times 128$, passed through 3 convolutional layers with maxpooling, followed by 3 fully connected layers. The output depth features have a dimension of 32.

For detecting contact, information from the force-torque sensor and the tactile sensors are significant. The 6 force-torque values are obtained from the 6 DoF Force-Torque sensor on the Barrett hand. A $24 \times 4$ dimension tactile data is obtained from the tactile sensors, with 24 values for each finger and 24 values for the palm. These values are flattened to form a one-dimensional vector. The force-torque values, tactile values, and 7 joint position values are concatenated and passed through a 3 layer MLP to produce 12 features. Finally, these 12 features along with the features from the color and depth images are passed through a 2 layer MLP to predict a vector of 4 binary values $b = (b_1,b_2,b_3,b_4)$. A cross-entropy loss is used for training the network.

The first output $b_1$ represents the contact state, a binary decision on if there is contact or not. The second value $b_2$  indicates if the contact type is point or line contact. The third value $b_3$ indicates a contact on the finger tactile sensors. The final value $b_4$ is a decision on if the scene is occluded by the arm. The relationship between vision and force to detect contact enables inferring these conditions using visual and tactile data. The contact can be seen and felt on the force torque sensor when the robot touches the surface. Thus, contact information can be inferred by combining data from various sensors.

The training of this network is performed using self-supervised methods to avoid manual labelling. The training is carried out by using both a random policy and a heuristic policy. The heuristic policy drives the robot to establish contact in directions seen in the demonstrations. The four outputs serve as input to the skill policy. 

\subsection{Skill Policy}
The policy for executing contact-rich skills is learnt using reinforcement learning. For contact-rich task the action space $A$ is not only kinematic, but can be forces as well. The  action directions where the force control is necessary is denoted by $A_F$. The directions perpendicular to this utilize position control and the action space for that is indicated by $A_x$.
\begin{equation}
    A = A_F \cup A_x
\end{equation}

The contact-rich skill learning is formulated as a model-free reinforcement learning problem. We use a maximum entropy RL algorithm as they tend to produce stable policies for practical RL applications \cite{avi2019rss}. Specifically, we use the soft actor-critic (SAC) algorithm to learn the policy. SAC uses a replay buffer $\mathcal{R}$ to store history and trains a critic and actor sampled from $\mathcal{R}$. The SAC algorithm is presented in Algorithm \ref{alg:sac}. The policy network used in MLP with input from the multimodal representation and produces actions -- position for $A_x$ and force for $A_F$.\\
\begin{algorithm}[h]
\SetAlgoLined
 Initialize policy $\pi$, critic $Q$, replay buffer $\mathcal{R}$\;
 \For{$i = 1 \to$ num\_iterations}{
 \For{ each step}{
 Sample $a_t$ from $\pi_\theta(a_t|s_t)$ and execute $a_t$\;
 Observe new state $s_{t+1}$\;
 Store $(s_t, a_t, r(s_t,a_t),s_{t+1})$ in $\mathcal{R}$\; 
 }
 \For{each gradient step}{
 Sample from $\mathcal{R}$\;
 Update policy $\pi$ and $Q$
 }
 }
 \caption{Soft actor-critic algorithm}
 \label{alg:sac}
\end{algorithm}
\textbf{Reward description} 
RL policies require designing reward functions which limit their applicability to complicated manipulation tasks. This is especially true for contact-rich manipulation tasks where designing the reward is difficult. One alternative is to learn the reward from the given demonstrations of the skill. The demonstrations contain information regarding goal of the skill which can be leveraged to learn the reward function. We propose to extract frames the demonstrations which represent positive and negative examples of goals and use them to train a classifier to represent the reward. Let $\mathcal{D}={(I_n,y_n)}$ denote the dataset of goal image frames $I_n$ and corresponding binary labels $y_n$ indicating positive or negative goals. The positive goal imply actual goal state, negative goal mean a failed state. Section \ref{sec:seg} describes how the positive goal frames are obtained. To obtain the negative examples, frames from the beginning of the demonstrations are used. Frames are randomly sampled from the first 30\% of the video before the second PIK. We make a reasonable assumption that the skill is not completed in beginning of the demonstration.
\begin{algorithm}[ht]
\SetAlgoLined
 \textbf{Provided:} $\mathcal{D}={(I_n,y_n)}$\;
Classifier $\nu(I)$ to minimize loss $\sum_n\mathcal{L}(\nu(I_n),y_n)$\\
Learn policy using reward $\phi(I_t)$ where,
\begin{equation}
    \phi(I_t,a_t) = \begin{cases} \nu(I_t)-1 & \text{if~} \nu(I_t) < \kappa\\
    1 & \text{if~} \nu(I_t)>\kappa 
    \end{cases}
\end{equation}
 \caption{Learning rewards from demonstration}
 \label{algo:rew}
\end{algorithm}
Classifier based rewards can enable learning tasks without reward engineering \cite{avi2019rss, xie2018few}. Let the binary classifier be denoted as $\nu(I)$. The goal classifier is trained as a network with a Resnet50 backend to extract image features and finally a 2 layer MLP with cross-entropy loss as the binary classification loss $\mathcal{L}$. Training the classifier requires a large number of examples of negative examples to avoid policy to fool the classifier by reaching states different from those shown to classifier. As the user provides multiple demonstration videos, large number of frames can be extracted to generate negative examples. The reward function is constructed based on the binary classification label as shown in Algorithm \ref{algo:rew}. $I_t$ is the image at time step $t$, and $\kappa$ is a threshold to provide high sparse reward when the performance is close to demonstration.
\begin{figure}[t]
    \centering
    \includegraphics[width=0.47\textwidth]{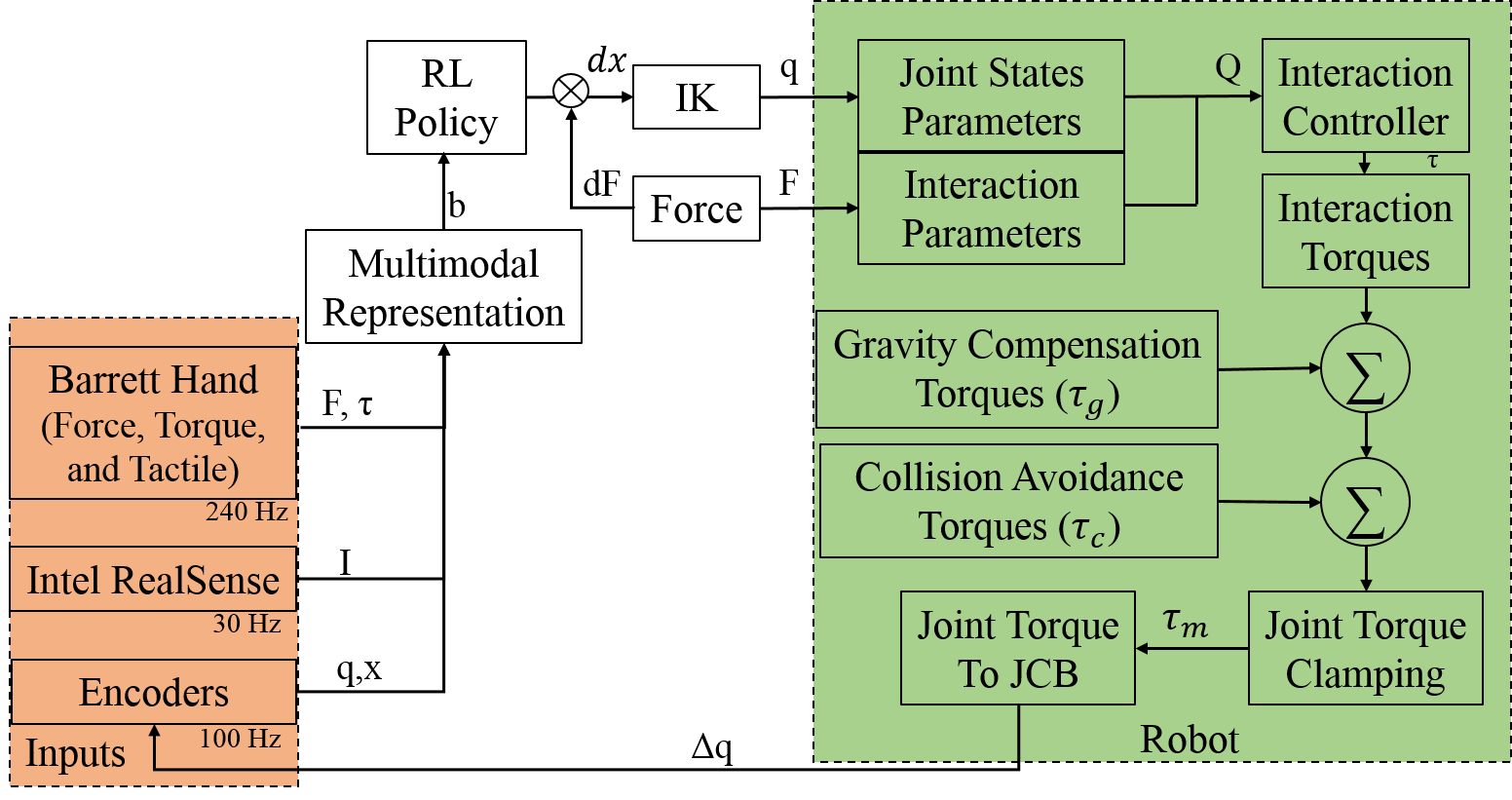}
    \caption{Execution Controller Model. Here $Q$ is the controller input and gain parameters, $\tau_m$ is the joint torques after clamping to motor limits.}
    \label{fig:controller}
\end{figure}

\subsection{The Robot Execution Controller}
Learning contact-rich policies require both position and force control modes. The controller uses the output position offsets $dx$ and force offsets $df$  provided from the policy and employs a hybrid control scheme to set the joint torques. The controller layout is shown in Fig. \ref{fig:controller}. The inputs to the controller is position and force offsets in Cartesian frame. Let $x$ be the current pose in 6 DoF  and current force along $A_F$ direction as $F$. $x_{t+1} = x_t + dx$ and $F_{t+1} = F_t + dF$. Let $X_d$ be the full 6 DoF Cartesian pose required and $X$ be the current pose. The required joint positions, $q_d$, can be computed using inverse kinematics. The impedance controller is used for all directions expect the force direction. In the direction of $F$, we use a force controller defined as follows,
\begin{equation}
    F_d = K(X_d - X) + C(\dot{X}_d - \dot{X})
\end{equation}
where $F_d$ is the impedance forces, $K$ is the stiffness and $C$ is the damping. Let $\tau_{ns}$ be the desired nullspace torques and $q$ the current joint angles. The required joint torques can be computed as $\tau = J^T(q)F + \tau_{ns}$.

%% file: experiments.tex
\section{Experiments and Results}
The experiments to evaluate and validate the performance of the skill modeling framework have been conducted for a) writing, b) cleaning, and c) peeling skills. All three are everyday contact-rich tasks requiring both vision and force feedback, particularly useful in robotic applications.

Two baselines were implemented to compare the performance of the proposed multi-modal skill learning framework. The first method is based on modeled controllers, derived from the work presented in \cite{balakuntala2019extending}. The second method is a naive agent that uses an RL agent with an engineered reward. The parameters for the modeled controller and the reward for the naive agent are designed based on the task. These rewards are described in detail in respective section.

\subsection{Hardware Setup}
\begin{figure}[t]
    \centering
    \includegraphics[width=0.4\textwidth]{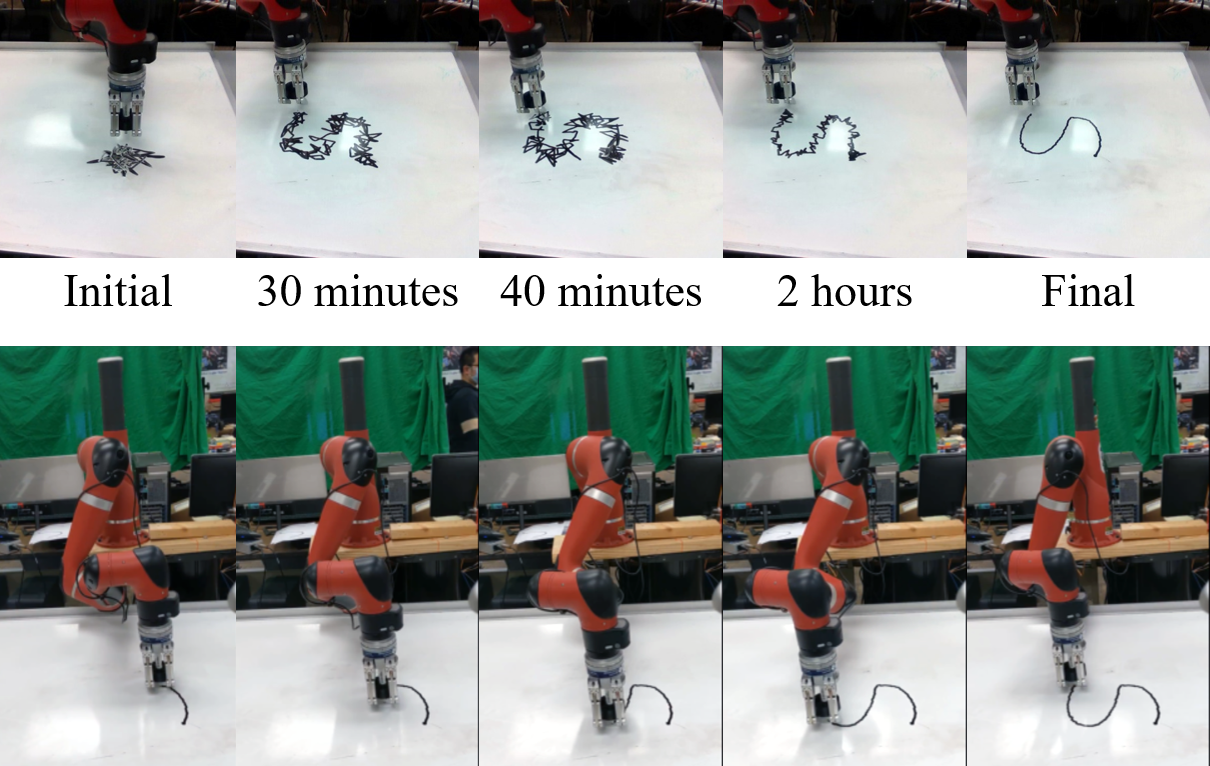}
    \caption{The results of writing the character `S'. The top image shows the results at different stages of training. The bottom figure shows a successful trial.}
    \label{fig:writing}
\end{figure}

Our experiments were conducted using the Rethink Sawyer robotic arm  which is widely used in robotics research, especially collaborative robotics. The entire setup is shown in Fig. \ref{fig:setup}. The robot is equipped with a Barrett hand gripper. The robot is controlled using impedance control mode. The Barrett BH-282 hand has tactile sensors on the three fingers and the palm, with a 6 DOF Force-Torque sensor at the wrist. An Intel Realsense D435i \cite{intel} is used for providing the color (RGB) and depth image frames. For all the experiments a region of interest is provided to reduce the background and concentrate focus on the workspace. The robot is also restricted to operate in a subset of the workspace in front of it and within the bounds of the region of interest. The same workspace bounds are used for all the tasks and this region can be seen in the top image in Fig. \ref{fig:writing}.
\subsection{Experiment 1: Writing}
In this experiment the goal is to write the demonstrated character. A total of 20 demonstrations are provided to the robot with the user writing the character `S'. The policy is trained using soft-actor critic with a 2 layer multilayer perceptron (MLP) network of size 64 which was trained for 40000 time steps which is approximately 6 hours of real world time.

The modeled controller uses the user hand trajectory from the demonstrations as the desired trajectory. For the naive RL agent, we use the total pixel difference between the mean demonstrated image and observed image as the penalty. If they overlap completely the reward will be 0 else the robot is penalized. The same policy is used for the execution. For the framework, when the occlusion parameter is 1, the reward $\phi(I_t,a_t)$ is set to 0. The threshold is set as, $\kappa = 0.8$.

The metric for task success is the amount of overlap between the two trajectories. A mean trajectory is computed from the demonstrated contours with a standard deviation, across the demonstrations. If the robot achieves a trajectory within 2 standard deviations from the mean demonstrated trajectory the execution is considered successful. Fig. \ref{fig:writing} shows the performance at different stages of training and the bottom image in the figure shows a successful trial. Notably, the framework achieves a success rate of 100\% after just 6 hours of training. The modeled controller achieves a success rate of 78\% and the baseline controller has a success rate of 30\% after 6 hours as shown in Table \ref{tab:task_success}.
\begin{figure}[t]
    \centering
    \includegraphics[width=0.4\textwidth]{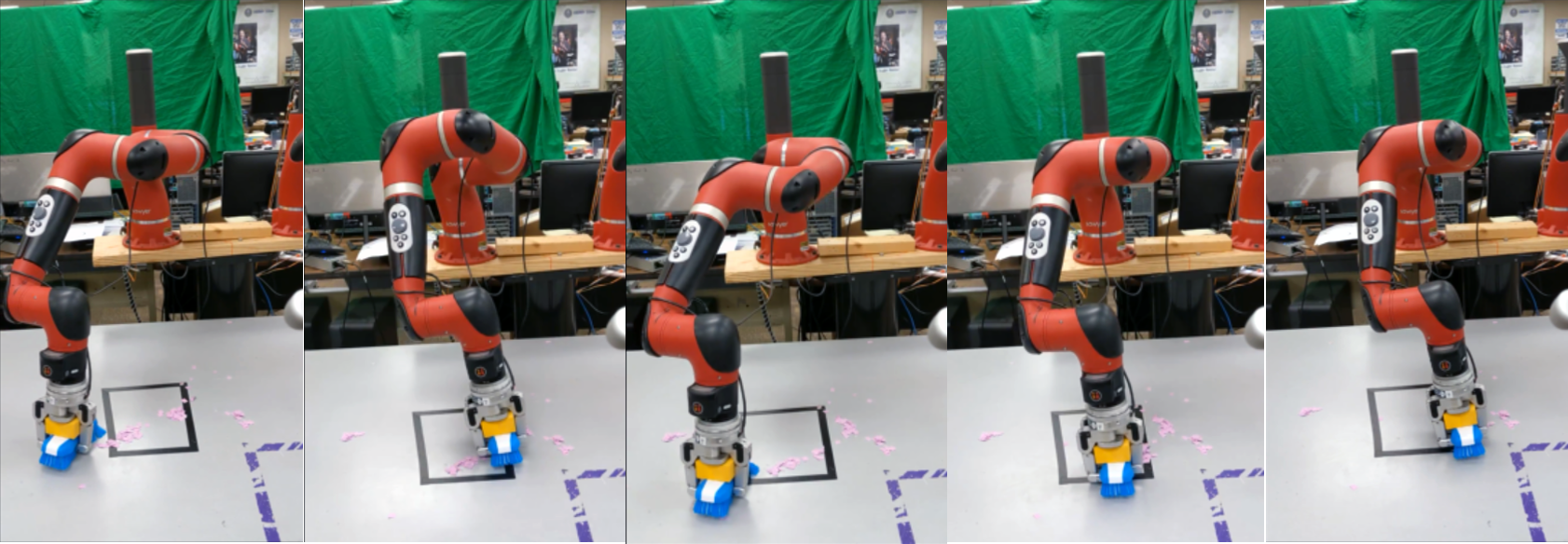}
    \caption{The robot performing the cleaning skill. The sequence of images shows a successful trial.}
    \label{fig:cleaning}
\end{figure}
\subsection{Experiment 2: Cleaning}
The second experiment is learning a surface cleaning task using a brush. Small bits of paper were scattered in the region of interest with the goal to clean the entire area inside this region. A total of 40 demonstrations were provided, with varying distribution of the particles as the demonstration data. The same policy, as described previously, was used for training using a soft-actor critic with a 2 layer MLP network of size 64. The policy was trained for 50000-time steps which are approximately 8 hours in real-world time.

The modeled controller approach uses the hand trajectory of the hand of the user from the demonstration as a feedback. For designing the reward for the naive RL agent, the area of the particles is used. The particles are detected in the RGB image using color thresholding methods. The ratio of the area of the particles to that of the region of interest in the image is used as the penalty. If $S_p$ is area of particles in image and $S_{ROI}$ is the area of the region of interest, then the reward ($r$) is calculated as: $r = -S_p/S_{ROI}$. As before, for the proposed skill learning framework, the reward $\phi(I_t,a_t)$ is set to 0 when the occlusion parameter is 1. The threshold $\kappa$ is set as: $\kappa = 0.8$.

The task success metric is based on the change in the area cleaned. With the final area ($S^f_p$), the task is considered successful if $S^f_p/S_{ROI} < 0.05$, i.e. 95\% of the surface is cleaned. Notably, the proposed framework achieves a success rate of 80\% with the modelled controller achieving a success rate of 30\% and the naive RL agent has a success rate of 43\% as shown in Table \ref{tab:task_success}.  Fig. \ref{fig:cleaning} shows the robot performing a successful trial using the policy from the proposed framework.

\subsection{Experiment 3: Peeling}
Peeling is a unique contact-rich task that requires the robot to interact with the object directly instead of a surface. The state change is observed on the object through vision in the case of peeling a cucumber. The goal of the experiment is to peel an area of the cucumber. 
A total of 30 demonstrations of a user peeling a side of a cucumber were provided to the robot. Using the same policy as described previously, the trained was accomplished for a soft-actor critic with a 2 layer MLP network of size 64. The policy was trained for 1000 time steps which are approximately equal to 30 minutes of real-world time.

The modeled controller approach extracts the the trajectory of the hand from the demonstrations as training input. Force bounds are specified to not to damage the object. For the naive RL agent, a image based reward is provided similar to the previous task. The exploration space is limited to the region of interest. Color thresholding is used to detect the area of the peeled cucumber and the unpeeled area. The ratio of the areas is provided as the reward. As before, for the proposed skill learning framework the reward $\phi(I_t,a_t)$ is set to 0 when the occlusion parameter is 1. The threshold $\kappa$ is set as: $\kappa = 0.8$.
\begin{figure}[t]
    \centering
    \includegraphics[width=0.4\textwidth]{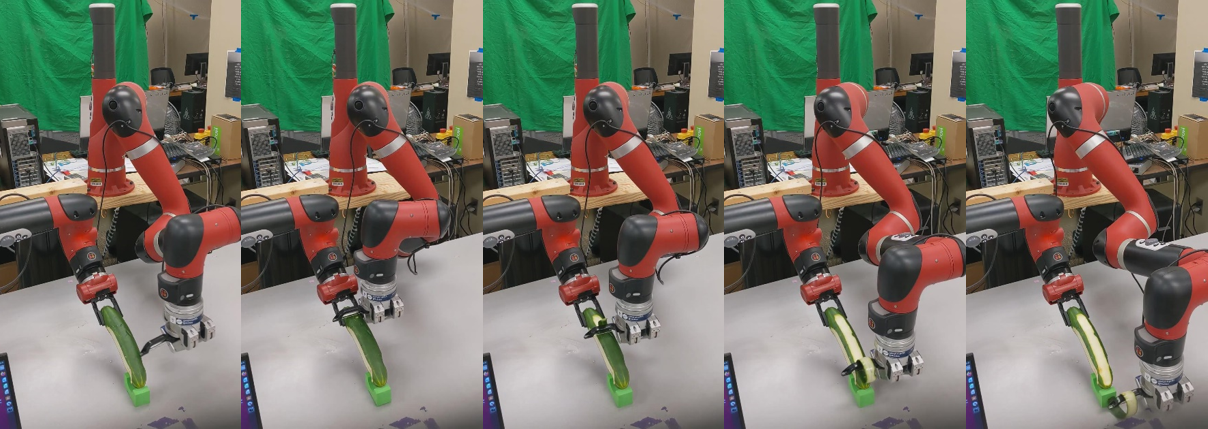}
    \caption{A successful trial of robot performing the peeling skill.}
    \label{fig:peeling}
\end{figure}
Instance segmentation is performed on the scene to detect the unpeeled and peeled areas of cucumber. The average ratio of peeled to unpeeled areas from the demonstration is used as the reference. If the ratio for the robot execution is within 10\% of the reference, the trial is considered successful. 

The proposed framework achieves a success rate of 100\%, the modeled controller achieves a success rate of 65\%, and the naive RL agent has a success rate 15\%. One of the successful trials using the skill learning framework is shown in Fig. \ref{fig:peeling}. Table \ref{tab:task_success} shows the task performance for the three methods. 
\begin{table}[ht]
    \centering
    \caption{Results on the success rate for the three contact-rich skills}
    \begin{tabular}{cccc}
        \toprule Skill & Proposed Method & Modeled controller & Naive RL \\\midrule
        Writing & 100\% & 78\% & 30\% \\
        Cleaning & 80\% & 30\% & 43\% \\
        Peeling & 100\% & 65\% & 15\% \\\bottomrule
    \end{tabular}
    \label{tab:task_success}
\end{table}

%% file: root.bbl
\begin{thebibliography}{10}
\providecommand{\url}[1]{#1}
\csname url@rmstyle\endcsname
\providecommand{\newblock}{\relax}
\providecommand{\bibinfo}[2]{#2}
\providecommand\BIBentrySTDinterwordspacing{\spaceskip=0pt\relax}
\providecommand\BIBentryALTinterwordstretchfactor{4}
\providecommand\BIBentryALTinterwordspacing{\spaceskip=\fontdimen2\font plus
\BIBentryALTinterwordstretchfactor\fontdimen3\font minus
  \fontdimen4\font\relax}
\providecommand\BIBforeignlanguage[2]{{%
\expandafter\ifx\csname l@#1\endcsname\relax
\typeout{** WARNING: IEEEtran.bst: No hyphenation pattern has been}%
\typeout{** loaded for the language `#1'. Using the pattern for}%
\typeout{** the default language instead.}%
\else
\language=\csname l@#1\endcsname
\fi
#2}}

\bibitem{blake2004neural}
R.~Blake, K.~V. Sobel, and T.~W. James, ``Neural synergy between kinetic vision
  and touch,'' \emph{Psychological science}, vol.~15, no.~6, pp. 397--402,
  2004.

\bibitem{song2014automated}
H.-C. Song, Y.-L. Kim, and J.-B. Song, ``Automated guidance of peg-in-hole
  assembly tasks for complex-shaped parts,'' in \emph{2014 IEEE/RSJ
  International Conference on Intelligent Robots and Systems}.\hskip 1em plus
  0.5em minus 0.4em\relax IEEE, 2014, pp. 4517--4522.

\bibitem{jmromano}
J.~M. {Romano}, K.~{Hsiao}, G.~{Niemeyer}, S.~{Chitta}, and K.~J.
  {Kuchenbecker}, ``Human-inspired robotic grasp control with tactile
  sensing,'' \emph{IEEE Transactions on Robotics}, vol.~27, no.~6, pp.
  1067--1079, 2011.

\bibitem{levine2016end}
S.~Levine, C.~Finn, T.~Darrell, and P.~Abbeel, ``End-to-end training of deep
  visuomotor policies,'' \emph{The Journal of Machine Learning Research},
  vol.~17, no.~1, pp. 1334--1373, 2016.

\bibitem{argall2009survey}
B.~D. Argall, S.~Chernova, M.~Veloso, and B.~Browning, ``A survey of robot
  learning from demonstration,'' \emph{Robotics and autonomous systems},
  vol.~57, no.~5, pp. 469--483, 2009.

\bibitem{calinon2007teacher}
S.~Calinon and A.~G. Billard, ``What is the teacher’s role in robot
  programming by demonstration?: Toward benchmarks for improved learning,''
  \emph{Interaction Studies}, vol.~8, no.~3, pp. 441--464, 2007.

\bibitem{calinon2007incremental}
S.~Calinon and A.~Billard, ``Incremental learning of gestures by imitation in a
  humanoid robot,'' in \emph{Proceedings of the ACM/IEEE international
  conference on Human-robot interaction}.\hskip 1em plus 0.5em minus
  0.4em\relax ACM, 2007, pp. 255--262.

\bibitem{balakuntala2019extending}
M.~V. Balakuntala, V.~L. Venkatesh, J.~P. Bindu, R.~M. Voyles, and J.~Wachs,
  ``Extending policy from one-shot learning through coaching,'' in \emph{2019
  28th IEEE International Conference on Robot and Human Interactive
  Communication (RO-MAN)}.\hskip 1em plus 0.5em minus 0.4em\relax IEEE, 2019,
  pp. 1--7.

\bibitem{billard2016learning}
A.~G. Billard, S.~Calinon, and R.~Dillmann, ``Learning from humans,'' in
  \emph{Springer handbook of robotics}.\hskip 1em plus 0.5em minus 0.4em\relax
  Springer, 2016, pp. 1995--2014.

\bibitem{finn2017one}
C.~Finn, T.~Yu, T.~Zhang, P.~Abbeel, and S.~Levine, ``One-shot visual imitation
  learning via meta-learning,'' in \emph{Conference on Robot Learning}, 2017,
  pp. 357--368.

\bibitem{finn2017deep}
C.~Finn and S.~Levine, ``Deep visual foresight for planning robot motion,'' in
  \emph{2017 IEEE International Conference on Robotics and Automation
  (ICRA)}.\hskip 1em plus 0.5em minus 0.4em\relax IEEE, 2017, pp. 2786--2793.

\bibitem{abbeel2004apprenticeship}
P.~Abbeel and A.~Y. Ng, ``Apprenticeship learning via inverse reinforcement
  learning,'' in \emph{Proceedings of the twenty-first international conference
  on Machine learning}.\hskip 1em plus 0.5em minus 0.4em\relax ACM, 2004, p.~1.

\bibitem{avi2019rss}
A.~Singh, L.~Yang, K.~Hartikainen, C.~Finn, and S.~Levine, ``End-to-end robotic
  reinforcement learning without reward engineering,'' in \emph{Robotics:
  Science and Systems}, 2019.

\bibitem{lee2019making}
M.~A. Lee, Y.~Zhu, K.~Srinivasan, P.~Shah, S.~Savarese, L.~Fei-Fei, A.~Garg,
  and J.~Bohg, ``Making sense of vision and touch: Self-supervised learning of
  multimodal representations for contact-rich tasks,'' in \emph{2019
  International Conference on Robotics and Automation (ICRA)}.\hskip 1em plus
  0.5em minus 0.4em\relax IEEE, 2019, pp. 8943--8950.

\bibitem{chebotar2017path}
Y.~Chebotar, M.~Kalakrishnan, A.~Yahya, A.~Li, S.~Schaal, and S.~Levine, ``Path
  integral guided policy search,'' in \emph{2017 IEEE international conference
  on robotics and automation (ICRA)}.\hskip 1em plus 0.5em minus 0.4em\relax
  IEEE, 2017, pp. 3381--3388.

\bibitem{zhu2018reinforcement}
Y.~Zhu, Z.~Wang, J.~Merel, A.~Rusu, T.~Erez, S.~Cabi, S.~Tunyasuvunakool,
  J.~Kram{\'a}r, R.~Hadsell, N.~de~Freitas, \emph{et~al.}, ``Reinforcement and
  imitation learning for diverse visuomotor skills,'' \emph{arXiv preprint
  arXiv:1802.09564}, 2018.

\bibitem{kalakrishnan2011learning}
M.~Kalakrishnan, L.~Righetti, P.~Pastor, and S.~Schaal, ``Learning force
  control policies for compliant manipulation,'' in \emph{2011 IEEE/RSJ
  International Conference on Intelligent Robots and Systems}.\hskip 1em plus
  0.5em minus 0.4em\relax IEEE, 2011, pp. 4639--4644.

\bibitem{lee2015learning}
A.~X. Lee, H.~Lu, A.~Gupta, S.~Levine, and P.~Abbeel, ``Learning force-based
  manipulation of deformable objects from multiple demonstrations,'' in
  \emph{2015 IEEE International Conference on Robotics and Automation
  (ICRA)}.\hskip 1em plus 0.5em minus 0.4em\relax IEEE, 2015, pp. 177--184.

\bibitem{bekiroglu2011learning}
Y.~Bekiroglu, R.~Detry, and D.~Kragic, ``Learning tactile characterizations of
  object-and pose-specific grasps,'' in \emph{2011 IEEE/RSJ international
  conference on Intelligent Robots and Systems}.\hskip 1em plus 0.5em minus
  0.4em\relax IEEE, 2011, pp. 1554--1560.

\bibitem{calandra2018more}
R.~Calandra, A.~Owens, D.~Jayaraman, J.~Lin, W.~Yuan, J.~Malik, E.~H. Adelson,
  and S.~Levine, ``More than a feeling: Learning to grasp and regrasp using
  vision and touch,'' \emph{IEEE Robotics and Automation Letters}, vol.~3,
  no.~4, pp. 3300--3307, 2018.

\bibitem{watkins2019multi}
D.~Watkins-Valls, J.~Varley, and P.~Allen, ``Multi-modal geometric learning for
  grasping and manipulation,'' in \emph{2019 International conference on
  robotics and automation (ICRA)}.\hskip 1em plus 0.5em minus 0.4em\relax IEEE,
  2019, pp. 7339--7345.

\bibitem{kappler2015data}
D.~Kappler, P.~Pastor, M.~Kalakrishnan, M.~W{\"u}thrich, and S.~Schaal,
  ``Data-driven online decision making for autonomous manipulation.'' in
  \emph{Robotics: Science and Systems}, 2015.

\bibitem{abu2015adaptation}
F.~J. Abu-Dakka, B.~Nemec, J.~A. J{\o}rgensen, T.~R. Savarimuthu,
  N.~Kr{\"u}ger, and A.~Ude, ``Adaptation of manipulation skills in physical
  contact with the environment to reference force profiles,'' \emph{Autonomous
  Robots}, vol.~39, no.~2, pp. 199--217, 2015.

\bibitem{beltran}
C.~C. {Beltran-Hernandez}, D.~{Petit}, I.~G. {Ramirez-Alpizar}, T.~{Nishi},
  S.~{Kikuchi}, T.~{Matsubara}, and K.~{Harada}, ``Learning force control for
  contact-rich manipulation tasks with rigid position-controlled robots,''
  \emph{IEEE Robotics and Automation Letters}, vol.~5, no.~4, pp. 5709--5716,
  2020.

\bibitem{ijspeert2013dynamical}
A.~J. Ijspeert, J.~Nakanishi, H.~Hoffmann, P.~Pastor, and S.~Schaal,
  ``Dynamical movement primitives: learning attractor models for motor
  behaviors,'' \emph{Neural computation}, vol.~25, no.~2, pp. 328--373, 2013.

\bibitem{gkioxari2018detecting}
G.~Gkioxari, R.~Girshick, P.~Doll{\'a}r, and K.~He, ``Detecting and recognizing
  human-object interactions,'' in \emph{Proceedings of the IEEE Conference on
  Computer Vision and Pattern Recognition}, 2018, pp. 8359--8367.

\bibitem{redmon2018yolov3}
J.~Redmon and A.~Farhadi, ``Yolov3: An incremental improvement,'' \emph{arXiv
  preprint arXiv:1804.02767}, 2018.

\bibitem{he2017mask}
K.~He, G.~Gkioxari, P.~Doll{\'a}r, and R.~Girshick, ``Mask r-cnn,'' in
  \emph{Proceedings of the IEEE international conference on computer vision},
  2017, pp. 2961--2969.

\bibitem{lesort2018state}
T.~Lesort, N.~D{\'\i}az-Rodr{\'\i}guez, J.-F. Goudou, and D.~Filliat, ``State
  representation learning for control: An overview,'' \emph{Neural Networks},
  vol. 108, pp. 379--392, 2018.

\bibitem{he2016deep}
K.~He, X.~Zhang, S.~Ren, and J.~Sun, ``Deep residual learning for image
  recognition,'' in \emph{Proceedings of the IEEE conference on computer vision
  and pattern recognition}, 2016, pp. 770--778.

\bibitem{xie2018few}
A.~Xie, A.~Singh, S.~Levine, and C.~Finn, ``Few-shot goal inference for
  visuomotor learning and planning,'' in \emph{Conference on Robot Learning},
  2018, pp. 40--52.

\bibitem{intel}
\BIBentryALTinterwordspacing
``intel® realsense™ depth and tracking cameras,'' Aug 2020. [Online].
  Available: \url{https://software.intel.com/en-us/realsense/d400}
\BIBentrySTDinterwordspacing

\end{thebibliography}
